\documentclass[runningheads]{llncs}
\usepackage{graphicx}

\usepackage{tikz}
\usepackage{comment}
\usepackage{amsmath,amssymb} 
\usepackage{color}
\usepackage{epsfig}
\usepackage{graphicx}
\usepackage[font=small,labelfont=bf]{caption}
\usepackage{mathtools}
\usepackage{stmaryrd}
\usepackage{booktabs}
\usepackage{physics}
\usepackage{wrapfig}
\usepackage{hyperref}
\usepackage[accsupp]{axessibility} 

\makeatletter
\def\blfootnote{\gdef\@thefnmark{}\@footnotetext}
\makeatother

\newcommand{\superscript}[1]{\ensuremath{^{\textrm{#1}}}}

\begin{document}
\pagestyle{headings}
\mainmatter
\def\ECCVSubNumber{6589} 

\title{Physically Disentangled Representations}

\titlerunning{Physically Disentangled Representations}

\author{Tzofi Klinghoffer\superscript{* 1} \and
Kushagra Tiwary\superscript{* 1} \and
Arkadiusz Balata\superscript{1} \and\\
Vivek Sharma\superscript{1,2} \and
Ramesh Raskar\superscript{1}}
\authorrunning{T. Klinghoffer et al.}
%

\institute{\superscript{1}Massachusetts Institute of Technology, \superscript{2}Harvard Medical School \\
\email{\{tzofi,ktiwary,arkadius,vvsharma,raskar\}@mit.edu}
}
\maketitle
\begin{abstract}
State-of-the-art methods in generative representation learning yield semantic disentanglement, but typically do not consider physical scene parameters, such as geometry, albedo, lighting, or camera. We posit that inverse rendering, a way to reverse the rendering process to recover scene parameters from an image, can also be used to learn \emph{physically} disentangled representations of scenes without supervision. In this paper, we show the utility of inverse rendering in learning representations that yield improved accuracy on downstream clustering, linear classification, and segmentation tasks with the help of our novel Leave-One-Out, Cycle Contrastive loss (LOOCC), which improves disentanglement of scene parameters and robustness to out-of-distribution lighting and viewpoints. We perform a comparison of our method with other generative representation learning methods across a variety of downstream tasks, including face attribute classification, emotion recognition, identification, face segmentation, and car classification. Our physically disentangled representations yield higher accuracy than semantically disentangled alternatives across all tasks and by as much as 18\%. We hope that this work will motivate future research in applying advances in inverse rendering and 3D understanding to representation learning. Code is made available \underline{\href{https://github.com/tzofi/physically-disentangled-representations}{here}}.
\end{abstract}


\blfootnote{\superscript{*} Equal contribution.}


\begin{figure}[t] 
\begin{center}
   \includegraphics[width=0.8\linewidth]{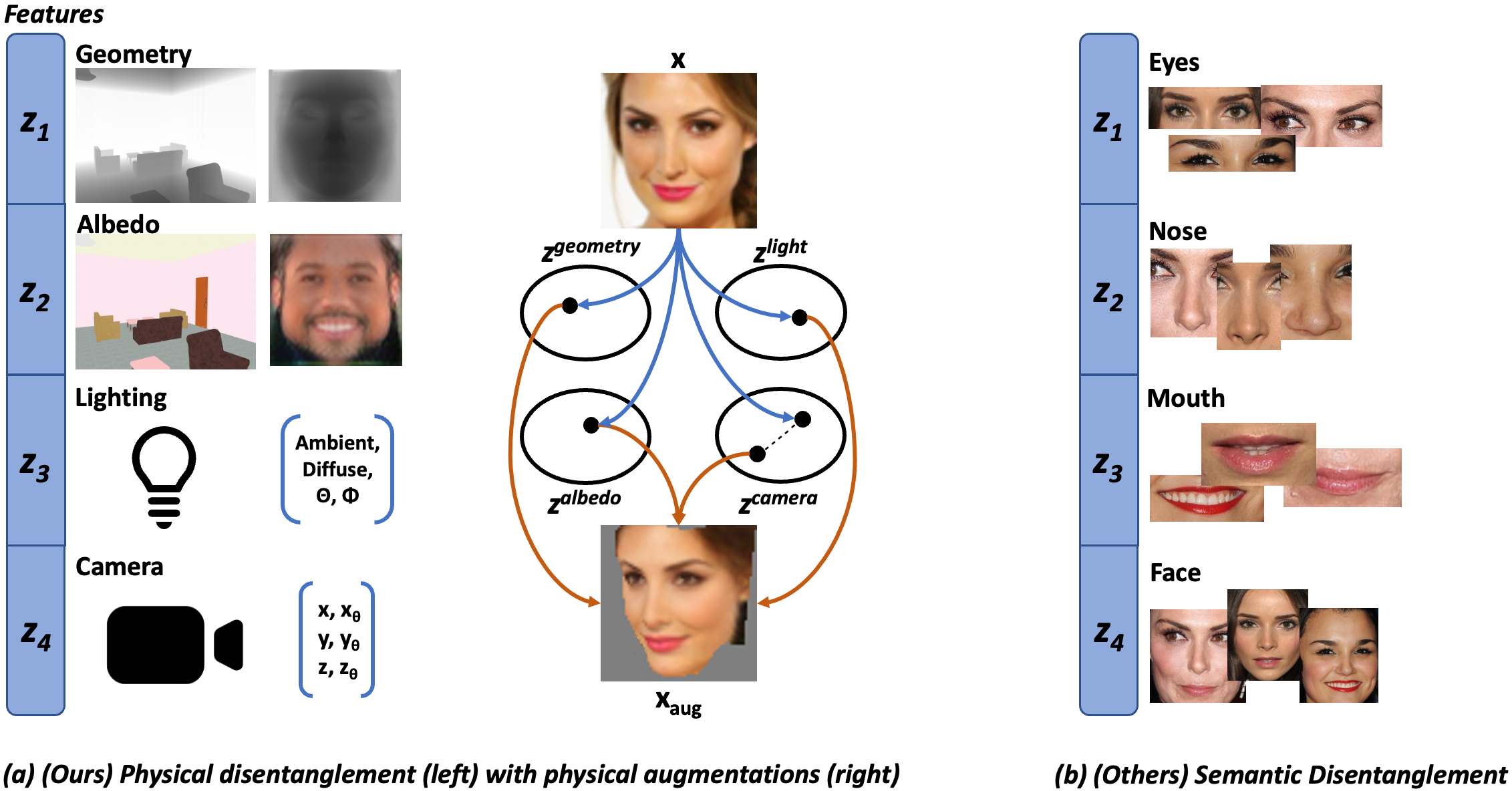}
\end{center}
\vspace{-4mm}
   \captionof{figure}{Overview: (a) we aim to achieve \emph{physical} disentanglement using a differentiable renderer in the loop during training, as opposed to (b) other generative representation learning methods, such as \cite{burgess2018understanding,chong2021retrieve,karras2020analyzing}, that learn semantically disentangled features.}
\label{figure:overview}
\vspace{-5mm}
\end{figure}

\vspace{-5mm}
\section{Introduction}


Unsupervised representation learning is a long-standing goal in the computer vision community. While representation learning has become widely popular in recent years as a way to mitigate dependencies on labeled data, learned features are typically said to capture semantic properties, not specific physical properties, such as geometry, albedo, lighting, or camera view. Existing representation learning methods can broadly be categorized as either generative or discriminative. Generative models rely on image generation or reconstruction as a signal for self-supervision, whereas discriminative models seek to predict known attributes of images as an auxiliary task. In the case of generative representation learning methods, such as as variational autoencoders (VAEs) \cite{kingma2013auto} and generative adversarial networks (GANs) \cite{goodfellow2014generative}, learned features have also been shown to be semantically disentangled \cite{bermano2022stateoftheart,burgess2018understanding}. Using a physical model, such as a renderer, to capture and disentangle the features of physical properties is less studied, but could be highly impactful. Motivated by the human ability to understand the world under different lighting conditions and viewpoints, we propose a method to learn and disentangle the features of physical properties using inverse rendering.

Inverse rendering provides a way for physical properties, also known as physical scene parameters, such as geometry, albedo, lighting, and camera view, to be predicted from images, and has rapidly advanced in recent years due to the availability of differentiable renderers. Whereas rendering, commonly used in graphics, synthesizes images from scene parameters, inverse rendering decomposes images into scene parameters. Because a differentiable renderer is used in training, features of complex physical parameters that are modeled by the renderer can be explicitly learned. By leveraging available priors, synthetic data, or implicit cues, such as symmetry, inverse rendering can be learned without supervision, and thus we show can be used for representation learning.

In this paper, we leverage existing inverse rendering methods to learn \emph{physically} disentangled representations that have utility across a wide-range of downstream tasks. We first introduce a general framework for representation learning using inverse rendering that can be adapted to fit existing inverse renderers. Since disentangling scene parameters without supervision is ill-posed, we then propose a novel loss term called Leave-One-Out, Cycle Contrastive loss (LOOCC) to improve disentanglement in the feature space. LOOCC applies contrastive learning to the features of images and their physically augmented counterparts, generated by a renderer. By generating augmented images with only a single changed scene parameter, we can then enforce that features of other scene parameters remain unchanged. Finally, we evaluate the learned representations on clustering, linear classification, and segmentation tasks. We show improved performance over strong baselines on all tasks, including face attribute classification, emotion recognition, identification, face segmentation, and car classification. Lastly, we discuss how including a renderer in learning can improve (1) interpretability by attributing predictions to physical scene parameters, and (2) robustness to physical phenomena, such as novel lighting and views, which remains a challenge in deep learning \cite{madan2020and,madan2021small}. In summary, we make the following contributions:

\begin{itemize}
  \setlength{\itemsep}{0mm}
  \item A general framework for learning physically disentangled feature representations through inverse rendering without supervision. Representations learned with our framework can effectively be used across many downstream tasks.
  \item A novel objective called Leave-One-Out, Cycle Contrastive loss (LOOCC) that helps disentangle the features of physical scene parameters such that they are more useful for downstream tasks.
  \item Detailed empirical results showing the benefit of our learned features for downstream clustering, linear classification, and segmentation tasks, including face attribute classification, emotion recognition, identification, face segmentation, and car classification.
  \item Discussions on the improvements the above contributions make to interpretability and robustness to out-of-distribution lighting and camera views.
\end{itemize}


\section{Related Work}

\subsection{Representation Learning}


Visual representation learning techniques aim to extract meaningful features from unlabeled images. Once models have been trained on the unlabeled data, learned representations can be transferred to downstream supervised learning tasks or used for applications such as image editing or synthesis.

\vspace{1mm}
\noindent
\textbf{Discriminative Methods}: Popular discriminative methods for representation learning include self-supervised learning and contrastive learning. Self-supervised learning methods use auxiliary tasks, such as predicting image rotation, to learn features from unlabeled data \cite{doersch2015unsupervised,doersch2017multi,komodakis2018unsupervised,noroozi2016unsupervised}. Contrastive learning methods, such as SimCLR \cite{chen2020simple}, minimize the mutual information between multiple views of an image, while retaining task-relevant features \cite{tian2020makes}. Typically, image views are created using standard data augmentation, rather than physical augmentations, such as changes to lighting or camera view, as done in our work. Learning disentangled features with discriminative representation learning methods remains an open problem \cite{burns2021unsupervised}. We use a renderer to disentangle features and create physical augmentations that are used in our contrastive loss.

\vspace{1mm}
\noindent
\textbf{Generative Methods}: VAEs \cite{burgess2018understanding,kingma2013auto,mathieu2019disentangling,van2017neural} and GANs \cite{bermano2022stateoftheart,nie2020semi} are commonly used for generative representation learning. Methods such as \cite{xia2021gan} utilize semantic disentanglement in the StyleGAN \cite{karras2019style,karras2020analyzing} latent space to perform image editing. Other methods, such as Retrieve in Style (RIS) \cite{chong2021retrieve}, further disentangle the StyleGAN2 latent code and use it for tasks such as image retrieval. Few papers have explored physical disentanglement in generative models. \cite{kulkarni2015deep} shows that the VAE latent space can be disentangled into graphics codes, such as pose, light, texture or shape, but do not utilize a renderer. \cite{wu2017neural} trains an encoder to learn disentangled scene parameters by using reinforcement learning with a non-differentiable graphics engine on synthetic data, while \cite{wu2017learning} extend this to real images of simple scenes. We use a differentiable renderer and demonstrate our method on images in the wild across many tasks.

\vspace{1mm}
\noindent
\textbf{3D Representations} Recent work has begun to explore how 3D information can be leveraged to improve the quality of learned representations. \cite{hou2021pri3d} proposes the use of an RGB-D dataset to render scenes from multiple views, using each view as an augmentation for a contrastive loss, and shows improved performance on downstream object detection and segmentation. \cite{kohli2020semantic} uses a scene representation network (SRN) \cite{sitzmann2019scene}, an implicit representation of a 3D scene, as a feature extractor for semi-supervised segmentation, training a segmentation classifier on top of the SRN features. Our work is motivated by these papers, and introduces a new way to learn physical representations that are useful for downstream tasks.

\vspace{-2mm}
\subsection{Inverse Rendering}

Neural inverse renderers use neural networks to predict scene parameters from images and differentiable renderers to perform image reconstruction. Wu \textit{et al.} \cite{wu2020unsupervised} accomplish unsupervised inverse rendering by leveraging symmetry in objects to create second views, but are thus constrained to highly symmetric objects such as human faces, cat faces, cars, etc. \cite{wu2021rendering} builds on this method by introducing a self-supervised discriminator to improve disentanglement of albedo in regions of high specularity, while \cite{wimbauer2022rendering} extends \cite{wu2020unsupervised} to non-symmetrical objects in the wild by including a geometry-prior in the form of a pre-trained depth network or structure from motion. \cite{li2020inverse,sengupta2019neural} both achieve inverse rendering of indoor scenes by first pre-training their models on synthetic data with ground truth for each scene parameter. \cite{yu2019inverserendernet} performs inverse rendering of outdoor scenes by utilizing multi-view stereo and a albedo consistency loss that enforces predicted albedo is consistent under different lighting. Our method utilizes the inverse rendering pipeline from \cite{wu2020unsupervised} for representation learning from unlabeled single views, assessing performance on downstream tasks, rather than image reconstruction.

\section{Proposed Method}

\begin{figure}[t] 
\begin{center}
   \includegraphics[width=1\linewidth]{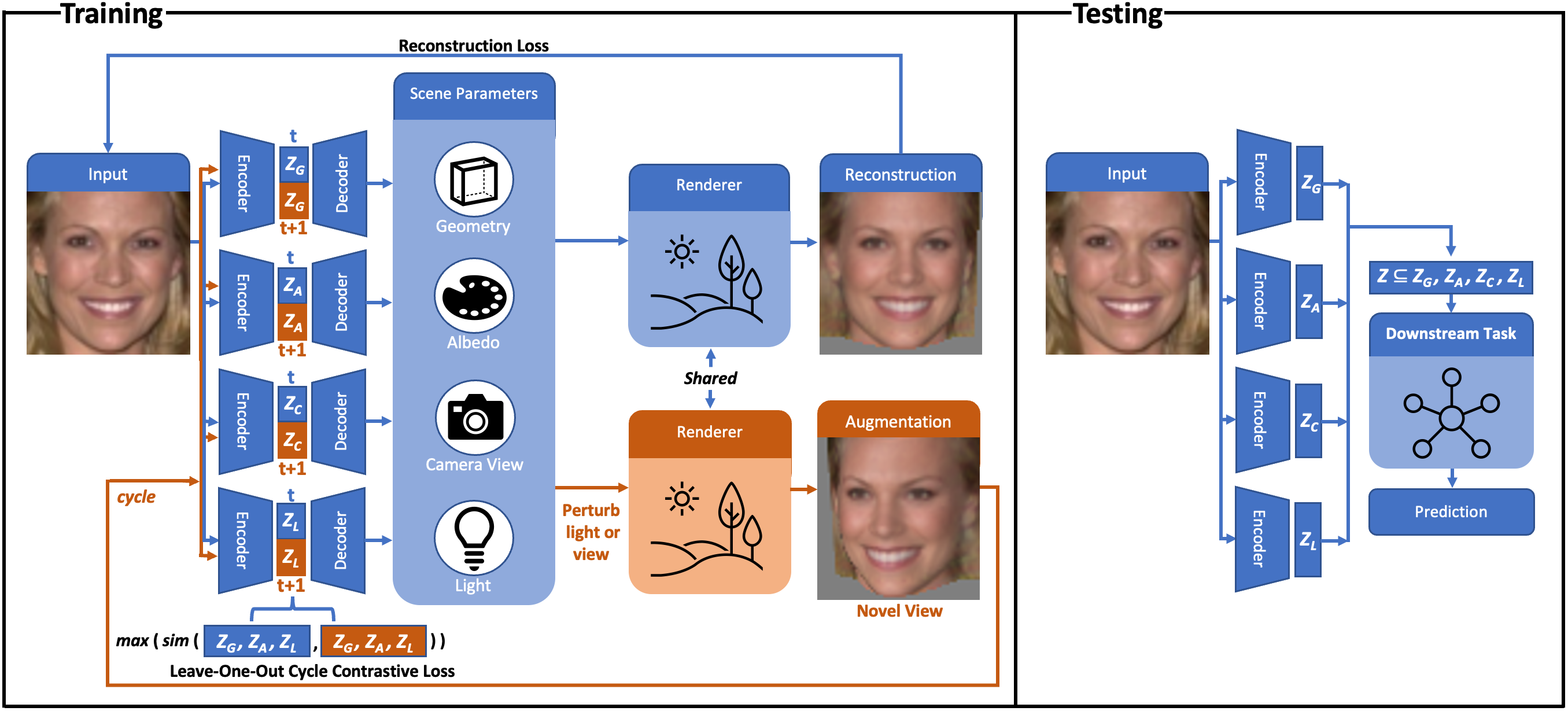}
\end{center}
\vspace{-4mm}
   \captionof{figure}{Our proposed method. Encoders are pre-trained for each physical scene parameter using an inverse rendering objective with a renderer in the loop. Learning is constrained by both a reconstruction loss and our novel Leave-One-Out, Cycle Contrastive (LOOCC) loss (illustrated in orange), requiring no supervision. LOOCC improves disentanglement between scene parameters by re-rendering images with new lighting or view, and enforcing that the features for other scene parameters remain similar. After training, the encoders can be used as feature extractors for downstream tasks.}
\label{figure:model}
\vspace{-2mm}
\end{figure}

In this section, we present our proposed method for learning representations with \emph{physical} disentanglement through the use of inverse rendering. First, we describe a general framework for representation learning through inverse rendering, and then we describe our proposed Leave-One-Out, Cycle Contrastive loss (LOOCC).

\subsection{Learning Representations with Inverse Rendering}
\label{sec:Inverse Rendering Framework}

We use inverse rendering as a mechanism to learn feature representations that are disentangled with regard to the physical scene parameters modeled by the renderer. Many existing inverse rendering methods use a shared encoder and separate decoders to estimate scene parameters \cite{li2020inverse,sengupta2019neural}. We instead propose the use of separate encoders per scene parameter to explicitly disentangle the scene parameter features. While this architectural modification can easily be implemented in methods that use a shared encoder, we instead utilize the work of Wu \textit{et al.} \cite{wu2020unsupervised}, which proposes a highly modular pipeline consisting of separate encoders and decoders for each scene parameter.

\vspace{2mm}
\noindent
\textbf{Neural Inverse Renderer:} The trainable inverse renderer is composed of a set of encoders $\mathcal{E_{\theta}}$, which each take an image \textbf{x} $\in \mathbb{R}^{H\times W\times3}$ to a physically disentangled feature, geometry, albedo, light, or camera. The output of each encoder is used to train a decoder $\mathcal{D_{\theta}}$ to predict the corresponding scene parameter. Predicted light values ($1 \times 4$) embed ambient and diffuse intensity, pitch, and yaw, while camera values embed camera rotation and translation in x, y, and z ($1 \times 6$). Both the encoders and decoders are parameterized by a neural network.

\begin{equation}
\begin{gathered} 
\label{encoder}
\mathcal{E_{\theta}}: \mathbb{R}^3 \xrightarrow[]{} \mathbb{R}^{1\times256} \\
(\textbf{x}) \mapsto (z^{geom}, z^{alb}, z^{cam}, z^{light}) \\
Z_{\textbf{x}} = \big ( z^{geom}, z^{alb}, z^{cam}, z^{light} \big ) 
\end{gathered} 
\end{equation}

\begin{equation}
\begin{gathered} 
\label{decoder}
\mathcal{D_{\theta}}: \mathbb{R}^{1\times256}\xrightarrow[]{} \mathbb{R}^{H\times W}\ \text{or}\
\mathbb{R}^{H\times W\times 3}\ \text{or}\
\mathbb{R}^{1\times4}\ \text{or}\ \mathbb{R}^{1\times6}\\ 
(z^{geom}, z^{alb}, z^{cam}, z^{light}) \mapsto (s_p^{geom}, s_p^{alb}, s_p^{cam}, s_p^{light}) \\ 
S_{p_{\textbf{x}}} = \big (s_p^{geom}, s_p^{alb}, s_p^{cam}, s_p^{light} \big )
\end{gathered} 
\end{equation}

\noindent
\textbf{Differentiable Renderer (NMR):} While $\mathcal{E_{\theta}}$ predicts physically disentangled features $Z_{\textbf{x}}$, their explicit counterparts, scene parameters $S_{p_{\textbf{x}}}$, are predicted by each decoder $\mathcal{D_{\theta}}$. The scene parameters are then fed into a differentiable renderer (NMR) \cite{kato2018neural}, $\mathcal{R}_{NMR}$. The NMR is responsible for constraining the encoder-decoders by reconstructing the input image \textbf{x}. 

\begin{equation}
\begin{gathered} 
\label{nmr_eq}
\mathcal{R}_{NMR}: \mathbb{R}^{H\times W} \times \mathbb{R}^{H\times W\times 3} \times \mathbb{R}^{1\times4} \times \mathbb{R}^{1\times6} \xrightarrow[]{} \mathbb{R}^{H\times W\times3} \\
(s_p^{geom}, s_p^{alb}, s_p^{cam}, s_p^{light}) \mapsto (\textbf{x}^{recon})
\end{gathered} 
\end{equation}


\noindent
\textbf{Representations:} Features are extracted from the last \emph{conv} layer of each encoder and stacked for use on downstream tasks. The stack can contain the features for all scene parameters (geometry, albedo, lighting, and camera view) or a subset. By including a subset, robustness to the omitted features can be improved, as decision-making no longer relies on the now omitted features, as discussed in Section 5. In this paper, we only utilize geometry and albedo features for downstream tasks because they capture the most information about the scene. We also observe that, surprisingly, light and camera features are useful on their own for many downstream tasks, as reported in the Appendix. In our method, we pre-train the model solely on the inverse rendering task without consideration of which downstream task(s) the features will be used for. Future work could harness the proposed framework to jointly train a learned representation with the target task in mind.


\subsection{Leave-One-Out, Cycle Contrastive Loss}

Disentangling scene parameters without supervision is a key challenge in inverse rendering. We propose a novel Leave-One-Out, Cycle Contrastive loss (LOOCC) to improve disentanglement as shown in Figure \ref{figure:overview}. Our method consists of physical augmentation, cyclic encoding, and contrastive learning.

\vspace{2mm}
\noindent
\textbf{Physical Augmentation:} In addition to reconstructing \textbf{x}, our method generates an augmented image of the scene by randomly perturbing a predicted scene parameter, $S_{p}$. Since the light and camera parameters are represented as four and six dimensional vectors, respectively, they can be perturbed by sampling a uniform distribution bounded by the desired range of each value. We randomly perturb light or camera while keeping other parameters the same, and use $\mathcal{R}_{NMR}$ to render the augmented image. We define a function $f$ that takes as input estimated scene parameters, $S_{p_\textbf{x}}$ for a given input image \textbf{x}, and randomly selects and perturbs either the light or camera parameter. In the below equation, we follow the example from Figure \ref{figure:model} and let $f$ perturb the camera parameter.

\begin{equation}
\begin{gathered} 
\label{nmr_eq}
f: \mathbb{R}^{H\times W} \times \mathbb{R}^{H\times W\times 3} \times \mathbb{R}^{1\times4} \times \mathbb{R}^{1\times6} \xrightarrow[]{} \\ \mathbb{R}^{H\times W} \times \mathbb{R}^{H\times W\times 3} \times \mathbb{R}^{1\times4} \times \mathbb{R}^{1\times6} \\
(s_{p_\textbf{x}}^{geom}, s_{p_\textbf{x}}^{alb}, s_{p_\textbf{x}}^{cam}, s_{p_\textbf{x}}^{light}) 
\mapsto 
(s_{p_{\textbf{x}}}^{geom}, s_{p_{\textbf{x}}}^{alb}, s_{p_{\textbf{x}_{aug}}}^{cam}, s_{p_{\textbf{x}}}^{light}) = S_{p_{\textbf{x}_{aug}}}
\end{gathered}
\end{equation}


\vspace{2mm}
\noindent
\textbf{Cyclic Encoding:} We leverage the observation that if we reconstruct an augmented image $\textbf{x}^{recon}_{aug}$ from $S_{p_{\textbf{x}_{aug}}}$, it should differ \emph{only} by one scene parameter. The augmented image can then cycle back through $\mathcal{E_{\theta}}$, generating a set of augmented features, $Z_{\textbf{x}_{aug}}$, that should be the same as the features of \textbf{x}, except for the features of the perturbed parameter. Following from the above example where the camera parameter is perturbed, we get $Z^U_{\textbf{x}_{aug}}$ which are the features of the unchanged scene parameters from the augmented image. 




\begin{equation}
\begin{gathered} 
\label{cyclic_rendering}
Z^U_{\textbf{x}_{aug}} = \big( z^{geom}_{\textbf{x}_{aug}}, z^{alb}_{\textbf{x}_{aug}}, z^{light}_{\textbf{x}_{aug}} \big) \subset \\ 
\hspace{40mm} \mathcal{E_{\theta}} \Big( \mathcal{R}_{NMR} \big(    
        s_{p_{\textbf{x}}}^{geom}, s_{p_{\textbf{x}}}^{alb}, s_{p_{\textbf{x}_{aug}}}^{cam}, s_{p_{\textbf{x}}}^{light}
        \big) \Big) \\ 
\end{gathered} 
\end{equation}

\vspace{2mm}
\noindent
\textbf{Leave-One-Out Contrastive Loss:} We leave out the features of the single perturbed scene parameter and use the contrastive loss proposed by \cite{chen2020simple} to enforce that the rest of the features in \textbf{x} and $\textbf{x}^{recon}_{aug}$ are similar, and thus that perturbing one scene parameter does not impact the features for the rest. Our intuition is that by leaving out one set of features, we allow these features in \textbf{x} and $\textbf{x}^{recon}_{aug}$ to be pushed apart, while the contrastive loss pulls the rest of the features together. Following the example from Figure \ref{figure:model} where camera view augmentation is shown, we arrive at the following equation for the normalized temperature-scaled cross entropy (contrastive) loss. We denote $Z^U_{\textbf{x}}$ to be all the features of the unchanged scene parameters from \textbf{x}, leaving out the features that were perturbed in $\textbf{x}^{recon}_{aug}$.


\begin{equation} \label{contrastive_loss}
L_{cont}(Z^U_{\textbf{x}}, Z^U_{\textbf{x}_{aug}}) = \frac{exp(Z^U_{\textbf{x}}, Z^U_{\textbf{x}_{aug}})}
{\sum_{k=1}^{2N} exp(sim(Z^U_{\textbf{x}},Z^{U^k}_{\textbf{x}})/\tau)}
\end{equation}

\noindent
In Equation \ref{contrastive_loss}, $sim$ measures cosine similarity of two feature vectors, $N$ is the minibatch size of the input, and $\tau$ is the temperature parameter.

\vspace{2mm}
\noindent
\textbf{Total Loss:} We define the total loss as a weighted sum of the reconstruction and LOOCC loss terms. In our work, we utilize the reconstruction loss proposed by Wu \emph{et al.} \cite{wu2020unsupervised}, but any reconstruction loss can be used.
\begin{equation} \label{total_loss}
\mathcal{L}(\textbf{x}) = \alpha \mathcal{L}_{cont} + \beta L_{recon}
\end{equation}

\noindent

While it is clear that modifying the lighting of a scene should not impact geometry, albedo, or camera, the impact of modifying the camera is less clear. While the geometry and albedo of a scene rendered from the perspective of two camera viewpoints will be different, we note that the encoder-decoders are tasked with recovering the canonical geometry and albedo. Thus, regardless of the camera view, the outputs of the encoder-decoders will be consistent, and we can safely enforce that changes in camera view do not result in changes in geometry or albedo. Finally, while the methods presented in this paper focus on augmenting light and camera view, they can also be extended to augment other scene parameters, assuming a sufficient sampling mechanism for perturbations. Since the current architecture uses autoencoders to estimate geometry and albedo, there is no explicit sampling method, but this could be overcome by using a VAE or network flow model to estimate geometry and albedo.




\section{Experiments}

\begin{figure}[t] 
\begin{center}
   \includegraphics[width=1\linewidth]{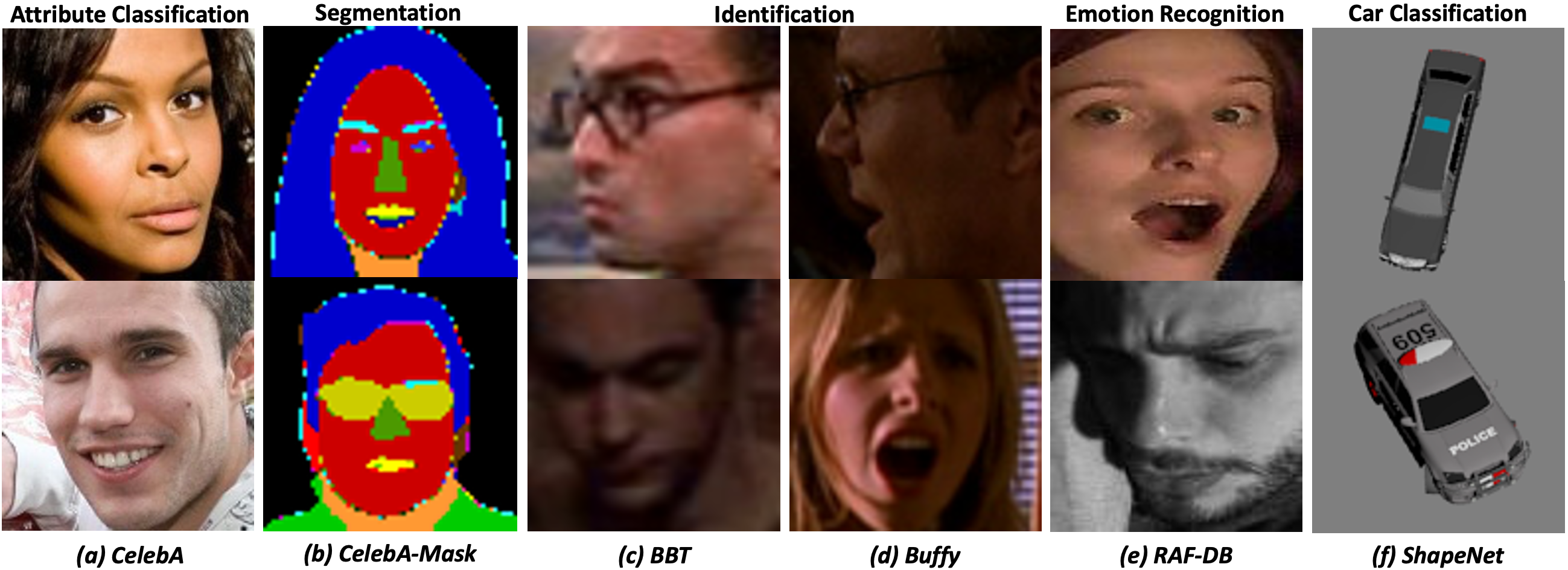}
\end{center}
\vspace{-5mm}
   \captionof{figure}{We demonstrate the utility of our method across a variety of tasks and challenging datasets, shown above. We achieve improved performance over strong baselines on all tasks. (b) shows segmentation results produced with our method.}
\label{figure:datasets}
\vspace{-6.5mm}
\end{figure}

In this section, we first introduce our datasets and evaluation metrics, followed by a description of our implementation. We then present the results of using the features learned by our method for many downstream tasks. We compare three versions of our method - No LOOCC, LOOCC with light augmentations (LOOCC-L), and LOOCC with light and view augmentations (LOOCC-LV).

\vspace{-2mm}
\subsection{Datasets}

\textbf{Training:} We train two sets of models with our proposed method. Our face models are trained on the UTK Face dataset \cite{zhifei2017cvpr}, containing 23,708 images, and we use an 80/10/10 split for train, validation, and test. Our car models are trained on the ShapeNet \cite{chang2015shapenet} cars dataset rendered by \cite{wu2020unsupervised}, consisting of 28k train images, 7k validation images, and 7k test images.

\vspace{1mm}
\noindent
\textbf{Testing:} We draw upon multiple datasets to test the performance of our method across different downstream tasks. We use the CelebA dataset \cite{liu2015faceattributes} for face attribute classification (Section 4.4), Buffy the Vampire Slayer (Buffy)~\cite{baeuml2013} and Big Bang Theory (BBT)~\cite{tapaswi2012} for identification (Section 4.5), the Real-world Affective Faces Database (RAF-DB) \cite{li2019reliable,li2017reliable} for emotion recognition (Section 4.6), the CelebA Mask dataset \cite{CelebAMask-HQ} for face segmentation (Section 4.7), and a subset of the ShapeNet cars test dataset mentioned above for car classification (Section 4.8). Each test dataset is described in more detail in its corresponding section.

\vspace{-2mm}
\subsection{Evaluation Metrics}

We evaluate our method on clustering, linear classification, and segmentation tasks. For clustering, we report hierarchical agglomerative clustering (HAC) as commonly used in past work in representation learning \cite{sharma2019self,sharma2020clustering,tapaswi2014total,zhang2016deep,zhang2016joint} with minimum variance ward linkage \cite{ward1963hierarchical}. To measure clustering performance, we report clustering accuracy (also known as weighted cluster purity) and F1-score. F1-score is computed using a weighted average. Clustering accuracy is computed by assigning the most common ground truth label for a cluster to all points in the cluster, as defined below.

\begin{equation} \label{cluster_accuracy_eqn}
Acc = \frac{1}{N}\sum_{c=1}^{\abs{C}} n_{c} \cdot p_{c}
\end{equation}

\noindent
In Equation \ref{cluster_accuracy_eqn}, $N$ is the number of samples, $n_{c}$ is the number of samples in cluster $c$, and $p_{c}$ is cluster purity, measured as the fraction of the largest number of samples from the same class to $n_{c}$. We set $C$ as the number of classes. For linear classification, we report accuracy using a threshold of 50\%. For segmentation, we report mean intersection over union (mIoU) and pixel accuracy.


\subsection{Implementation Details}

\begin{table*}[t]
\centering
\begin{tabular}{p{0.38\linewidth}p{0.17\linewidth}p{0.26\linewidth}p{0.15\linewidth}}
\hline
\textbf{Method} & \textbf{Dataset} & \textbf{Cluster Accuracy} & \textbf{F1-Score} \\
\hline
VQ-VAE (\texttt{NeuRIPS'17}) \cite{van2017neural} & BBT & 0.4168 & 0.2796 \\
StyleGAN2 (\texttt{CVPR'20}) \cite{karras2019style} & BBT & 0.4261 & 0.3528 \\
RIS (\texttt{ICCV '21}) \cite{chong2021retrieve} & BBT & 0.4790 & 0.4346 \\
Ours - No LOOCC & BBT & 0.5754 & 0.4572 \\
Ours - LOOCC-L & BBT & \textbf{0.6252} $\uparrow$  & \textbf{0.6133} $\uparrow$\\
Ours - LOOCC-LV & BBT & 0.6096 & 0.5669 \\
\hline
\hline
VQ-VAE (\texttt{NeuRIPS'17}) \cite{van2017neural} & Buffy & 0.4489 & 0.3427 \\
StyleGAN2 (\texttt{CVPR'20}) \cite{karras2019style} & Buffy & 0.3909 & 0.2767 \\
RIS (\texttt{ICCV '21}) \cite{chong2021retrieve} & Buffy & 0.3715 & 0.2837 \\
Ours - No LOOCC & Buffy & 0.4507 & 0.3697 \\
Ours - LOOCC-L & Buffy & 0.4718 & 0.4095 \\
Ours - LOOCC-LV & Buffy & \textbf{0.4806} $\uparrow$ & \textbf{0.4234} $\uparrow$ \\
\hline
\hline
VQ-VAE (\texttt{NeuRIPS'17}) \cite{van2017neural} & RAF-DB & 0.3993 & 0.2660 \\
StyleGAN2 (\texttt{CVPR'20}) \cite{karras2019style} & RAF-DB & 0.3863 & 0.2153 \\
RIS (\texttt{ICCV '21}) \cite{chong2021retrieve} & RAF-DB & 0.3862 & 0.2152 \\
Ours - No LOOCC & RAF-DB & 0.3918 & 0.2651 \\
Ours - LOOCC-L & RAF-DB & \textbf{0.4058} $\uparrow$ & \textbf{0.3014} $\uparrow$ \\
Ours - LOOCC-LV & RAF-DB & 0.396 & 0.2452 \\
\hline
\hline
VQ-VAE (\texttt{NeuRIPS'17}) \cite{van2017neural} & CelebA & 0.8010 & 0.7420 \\
StyleGAN2 (\texttt{CVPR'20}) \cite{karras2019style} & CelebA & 0.8042 & 0.7466 \\
RIS (\texttt{ICCV '21}) \cite{chong2021retrieve} & CelebA & 0.8023 &  0.7375 \\
Ours - No LOOCC & CelebA & 0.8135 &  0.7607 \\
Ours - LOOCC-L & CelebA & \textbf{0.8143} $\uparrow$ & 0.7616 \\
Ours - LOOCC-LV & CelebA & 0.8139 & \textbf{0.7649} $\uparrow$ \\
\hline
\end{tabular}
\captionof{table}{Clustering results for face classification tasks. Despite having a smaller latent space than the baselines, our method outperforms them, especially with the use of our novel Leave-One-Out, Cycle Contrastive loss (LOOCC). LOOCC-L indicates light augmentation and -LV indicates light and view augmentation. Tasks include identification (BBT, Buffy), emotion recognition (RAF-DB), and attribute classification (CelebA).}
\label{table:results_clustering}
\vspace{-10mm}
\end{table*}
Our implementation was developed in PyTorch 1.8.0 \cite{paszke2019pytorch}. Both our method and the VQ-VAE \cite{van2017neural} baselines are trained using the ADAM optimizer \cite{kingma2014adam} with an initial learning rate of $1\times10^{-3}$, and early stopping is used for both to prevent overfitting, with a patience of ten epochs. Models are trained and tested using $64\times64$ sized images. When the LOOCC loss is employed with our method, we use a temperature of 0.5 for Equation \ref{contrastive_loss}, and we set $\alpha$ to 0.01 and $\beta$ to 1.00 for Equation \ref{total_loss}. For light augmentation, we randomly perturb ambient light intensity, diffuse light intensity, and lighting direction by sampling uniform distributions between (-0.5, 0.5), (-0.5, 0.5), and (-45\textdegree, 45\textdegree), respectively, where the lighting direction is sampled for both pitch and yaw. For camera augmentation, we randomly perturb pitch and yaw by sampling uniform distributions between (-22.5\textdegree, 22.5\textdegree) and (-45\textdegree, 45\textdegree), respectively. All sampled values are added to the corresponding predicted parameter for the image. Unless otherwise stated, only geometry and albedo features are used for downstream tasks.

Our inverse rendering model has a total of 29,441,536 trainable parameters, with 12,979,648 for the encoder-decoders embedding the geometry and albedo features. Each scene parameter is represented by a $1 \times 256$ feature embedding, resulting in a $1\times512$ feature embedding for geometry and albedo together. We train the VQ-VAE with a latent dimension (D) of 416 and a latent dimension space (K) of 512, yielding 19,199,491 trainable parameters, and $416 \times 25 \times 25$ (260k) sized feature embeddings. We also trained the VQ-VAE model with multiple parameters and found the above parameters yielded the strongest baseline.

For all face classification tasks, we also compare our model against the state-of-the-art StyleGAN2 \cite{karras2020analyzing}, with \cite{tov2021designing} to invert images into the GAN latent space, and Retrieve In Style \cite{chong2021retrieve} (which uses a StyleGAN2 generator, again with \cite{tov2021designing} for inversion). These models are pre-trained on the large-scale Flickr-Faces-HQ Dataset (FFHQ), containing 70,000 training images (nearly 3x the number of training images used by our model), and are much larger at 297,506,764 trainable parameters in the generator alone (over 10x larger than our model, not including the discriminator). The feature embeddings of these models are also much larger than ours ($9216$ and $8 \times 9088$, respectively, where the 8 channels in the RIS embedding encode features for nose, eyes, mouth, hair, background, cheeks, neck, and clothes). For fair comparison to our model, we test each baseline with $64 \times 64$ sized images, but also observe that our model is competitive with baselines even when they are tested with $256 \times 256$ images, as reported in the Appendix.


\begin{table*}[t]
\centering
\begin{tabular}{p{0.35\linewidth}p{0.14\linewidth}p{0.25\linewidth}p{0.18\linewidth}}
\hline
\textbf{Method} & \textbf{\# Train} & \textbf{Acc (Finetune)} & \textbf{Acc (Frozen)} \\
\hline
Supervised ResNet-18 \cite{he2016deep} & 100 & 0.8186 & - \\
VQ-VAE (\texttt{NeuRIPS '17}) \cite{van2017neural} & 100 & 0.8148 & 0.8182 \\
StyleGAN2 (\texttt{CVPR '20}) \cite{karras2019style} & 100 & 0.7781 & 0.7688 \\
RIS (\texttt{ICCV '21}) \cite{chong2021retrieve} & 100 & 0.8250 & 0.8208 \\
Ours - No LOOCC & 100 & 0.8212 & 0.8033 \\
Ours - LOOCC-L & 100 & 0.8245 & 0.8205 \\
Ours - LOOCC-LV & 100 & \textbf{0.8556} $\uparrow$ & \textbf{0.8226} $\uparrow$ \\
\hline
\hline
Supervised ResNet-18 \cite{he2016deep} & 500 & 0.8336 & - \\
VQ-VAE (\texttt{NeuRIPS '17}) \cite{van2017neural} & 500 & 0.8349 & 0.8354 \\
StyleGAN2 (\texttt{CVPR '20}) \cite{karras2019style} & 500 & 0.8125 & 0.7875 \\
RIS (\texttt{ICCV '21}) \cite{chong2021retrieve} & 500 & 0.8550 & \textbf{0.8500} $\uparrow$ \\
Ours - No LOOCC & 500 & 0.8403 & 0.8166 \\
Ours - LOOCC-L & 500 & 0.8509 & 0.8219 \\
Ours - LOOCC-LV & 500 & \textbf{0.8556} $\uparrow$ & 0.8226 \\
\hline
\hline
Supervised ResNet-18 \cite{he2016deep} & 1000 & 0.8407 & - \\
VQ-VAE (\texttt{NeuRIPS '17}) \cite{van2017neural} & 1000 & 0.8368 & \textbf{0.8406} $\uparrow$ \\
StyleGAN2 (\texttt{CVPR '20}) \cite{karras2019style} & 1000 & 0.8063 & 0.7937 \\
RIS (ICCV '21) \cite{chong2021retrieve} & 1000 & 0.8250 & 0.8400 \\
Ours - No LOOCC & 1000 & 0.8539 & 0.8240 \\
Ours - LOOCC-L & 1000 & 0.8543 & 0.8226 \\
Ours - LOOCC-LV & 1000 & \textbf{0.8556} $\uparrow$ & 0.8149 \\
\hline
\end{tabular}
\captionof{table}{Linear classification accuracy for CelebA attribute classification with varying training samples. An MLP is trained atop each finetuned or frozen, pre-trained model.}
\label{table:results_semisupervised}
\vspace{-10mm}
\end{table*}

\subsection{Face Attribute Classification} 
\textbf{Details:} We use the representations learned by our method for face attribute classification, evaluated with both clustering and linear classification. Our test dataset consists of 6,000 randomly selected images from CelebA. For clustering, we extract the features from our model and run clustering 40 times for each of the binary attributes in CelebA, reporting the average. For linear classification, we train an MLP projection head (i.e. linear classifier) on top of the pre-trained model being evaluated, following \cite{chen2020simple}. We limit the labeled training samples to 100, 500, or 1000, and evaluate with the pre-trained model either frozen or fine-tuned, reporting each. Each classifier model is trained for 100 epochs.


\vspace{2mm}
\noindent
\textbf{Results:} Results for clustering and linear classification are reported in Table \ref{table:results_clustering} and Table \ref{table:results_semisupervised}, respectively. Across both forms of evaluation, our model outperforms the baselines, even as we vary the number of training samples for linear classification. Interestingly, fine-tuning our model seems to provide a consistent benefit, while fine-tuning the baselines does not. In analyzing the linear model, we observed that the two attributes that our method improved accuracy on most were high cheek bone and mouth slightly open, where we expect geometric features captured in our model to be most beneficial.

\vspace{-3mm}
\subsection{Face Identification}
\vspace{-1mm}
\textbf{Details:} We demonstrate the utility of our learned representations on face identification, specifically on the challenging task of video face clustering. We use Buffy season 5, episode 2 and BBT season 1, episode 1 as prepared by \cite{sharma2019self}. Both datasets contain extremely challenging out-of-distribution lighting and viewpoints, and thus test the robustness of our model. As in prior work \cite{sharma2017,sharma2019self}, clustering is done on a per-track basis by averaging the features of each frame in the track. Our Buffy dataset contains 568 tracks and six identities (Xander, Buffy, Dawn, Anya, Willow, Giles), and our BBT dataset contains 644 tracks and five identities (Howard, Leonard, Penny, Raj, Sheldon).

\vspace{2mm}
\noindent
\textbf{Results:} Our models outperform all baselines for clustering accuracy and F1 on both the Buffy and BBT datasets. Incorporation of the LOOCC loss is especially helpful since it improves disentanglement and thus robustness to lighting and viewpoints, which are challenging in both datasets. For BBT, the LOOCC loss with only light augmentations is best, whereas, for Buffy, the best clustering accuracy and F1 are achieved by the LOOCC model with light and view augmentations. During analysis, we observed that our models also have higher normalized mutual information (NMI) than all baselines on both Buffy and BBT. We achieve 43\% and 23\% on Buffy and BBT, respectively, whereas the best baseline methods, VQ-VAE and RIS, achieve 13\% and 18\%, respectively.


\vspace{-2mm}
\subsection{Emotion Recognition}
\textbf{Details:} We use clustering to perform emotion recognition using our learned representations. Our test dataset contains 3,068 images from RAF-DB, each containing one of the following seven emotions: surprise, fear, disgust, happiness, sadness, anger, and neutral.

\vspace{2mm}
\noindent
\textbf{Results:} Our method yields the strongest performance for emotion recognition over all metrics. We also observe that the LOOCC loss again yields improved performance over the no LOOCC alternative.

\vspace{-2mm}
\subsection{Face Segmentation}

\begin{table*}[h]
\vspace{-8mm}
\centering
\begin{tabular}{p{0.35\linewidth}p{0.15\linewidth}p{0.20\linewidth}p{0.25\linewidth}}
\hline
\textbf{Method} & \textbf{mIoU} & \textbf{Top 10 mIoU} & \textbf{Pixel Accuracy} \\
\hline
Supervised (random init) & 0.3862 & 0.5858 & 0.8669 \\
\hline
Geometry Encoder & 0.4066 & 0.5977 & 0.8715 \\
Albedo Encoder & 0.3987 &  0.5984 & 0.8724 \\
Joint Encoders & \textbf{0.4390} $\uparrow$ & \textbf{0.6221} $\uparrow$ & \textbf{0.8781} $\uparrow$ \\
Joint Encoders Frozen & 0.3962 & 0.5712 & 0.8540\\
\hline
\end{tabular}
\captionof{table}{U-net face segmentation results on the CelebA Mask dataset containing 19 classes. We report the mIoU, top 10 mIoU, and pixel accuracy, and compare the randomly initialized U-net with our method, which uses a pre-trained geometry or albedo U-net encoder and randomly initialized decoder. We also report performance when both the geometry and albedo encoders are jointly used as the U-net encoder. We allow the pre-trained encoders to be fine-tuned, except in the last row.}
\label{table:results_segmentation}
\vspace{-8mm}
\end{table*}

\noindent
\textbf{Details:} We test our learned features on segmentation using 24,127 training images and 2,885 test images from CelebA Mask, all downsampled to $64\times64$. Segmentation is done over 19 classes, containing instances such as eyes, mouth, etc. We leverage the U-net model \cite{ronneberger2015u} to perform segmentation and compare the supervised U-net with a U-net containing our pre-trained encoder, for either geometry or albedo. For pre-training, we replace and train $\mathcal{E_{\theta}}$ in our model with U-net encoders. We also report accuracy when both the geometry and albedo encoders are jointly used in U-net by stacking the features used for each skip connection, and experiment with both fine-tuned and frozen encoders. All segmentation models are trained for 20 epochs. Results are reported in Table \ref{table:results_segmentation}.

\vspace{2mm}
\noindent
\textbf{Results:} We observe that pre-training the U-net encoder with our method yields improvements in segmentation, both for mIoU and pixel accuracy. When both geometry and albedo encoders are jointly used and kept frozen, our model still outperforms supervised U-net in mIoU, despite fewer trainable parameters.


\vspace{-2mm}
\subsection{ShapeNet Car Classification}
\begin{table*}[t]
\centering
\begin{tabular}{p{0.40\linewidth}p{0.26\linewidth}p{0.16\linewidth}}
\hline
\textbf{Method} & \textbf{Cluster Accuracy} & \textbf{F1-Score} \\
\hline
VQ-VAE (NeuRIPS '17) \cite{van2017neural} & 0.4915 & 0.3696 \\
\hline \hline
Ours - No LOOCC & 0.5100 & 0.3877 \\
Ours - LOOCC-L &  0.5270 & 0.4016 \\
Ours - LOOCC-LV & \textbf{0.5485} $\uparrow$ & \textbf{0.4995} $\uparrow$ \\
\hline
\end{tabular}
\captionof{table}{Clustering results for ShapeNet car classification. Images contains either a police car, ambulance, limousine, jeep, or Ferrari (5 clusters).}
\label{table:results_car_clustering}
\vspace{-8mm}
\end{table*}

\noindent
\textbf{Details:} We utilize a subset of the ShapeNet car test data rendered by \cite{wu2020unsupervised} to evaluate car classification. For each image, we extract the car name from the ShapeNet metadata. Since ShapeNet contains a diverse set of cars, we limit our test set to five classes, using keyword matching to gather labels. The five classes are police car, ambulance, limousine, jeep, and Ferrari. The final test set contains 1000 images, each rendered with random lighting and viewpoint.

\vspace{2mm}
\noindent
\textbf{Results:} We compare our proposed method with VQ-VAE. Despite the VQ-VAE model having a much larger latent space, our method yields higher clustering accuracy and F1. As with results on face data, we observe that the inclusion of the LOOCC loss significantly improves performance, with light and view augmentation yielding the best performance.

\vspace{-2mm}
\section{Discussion}
\vspace{-2mm}
Our proposed method introduces inverse rendering as a framework for representation learning, and shows the utility of \emph{physically} disentangled representations on many downstream tasks. Although the majority of our benchmarked tasks are on face datasets, the introduced methods will scale across domains as inverse rendering continues to advance. While testing our model, we also discovered that the learned features are useful regardless of clustering method, and include results on FINCH \cite{finch}, which discovers a hierarchy of partitions without being given the number of clusters, in the Appendix. Beyond improved accuracy on downstream tasks, we posit that our framework has two other main advantages over existing methods: improved interpretability and improved robustness to physical phenomena that are both modeled by the renderer and disentangled.

\begin{figure}[t] 
\begin{center}
   \includegraphics[width=0.5\linewidth]{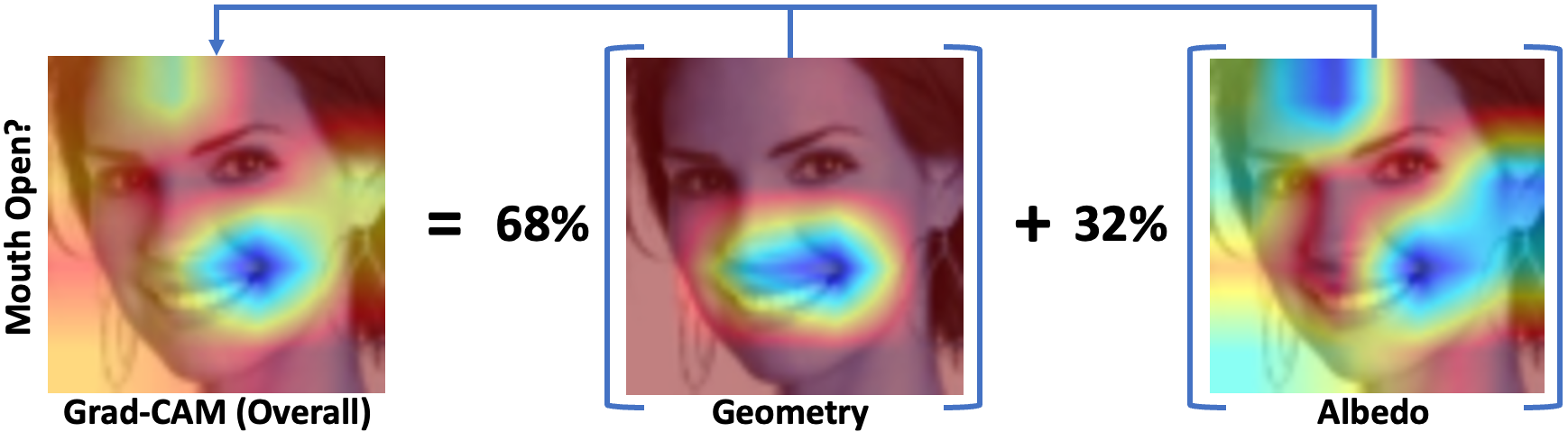}
\end{center}
\vspace{-6mm}
   \captionof{figure}{Interpretability of our model using Grad-CAM \cite{selvaraju2017grad} on the features of each scene parameter, with \% contribution of each feature computed as the normalized sum of absolute integrated gradients (IG) \cite{sundararajan2017axiomatic} for the feature: $\frac{\sum_{i=0}^{N} \abs{\text{IG}(z_{i}^{j})}}{\sum_{j=0}^{n}\sum_{i=0}^{N} \abs{\text{IG}(z_{i}^{j})}}$, where $z^{j}$ is the $N$-dimensional feature for scene parameter $j$. Blue indicates higher attribution.}
\label{figure:gradcam}
\vspace{-6mm}
\end{figure}

\vspace{1mm}
\noindent
\textbf{Applications to Interpretability:} Interpretability of deep learning models is a concern in many domains. While gradient-based attribution methods are a common way to gain high-level insight into model predictions, the resulting saliency maps often lack the granularity needed for sufficient model understanding. Such saliency maps highlight which pixels were most important for a model prediction. While this is helpful, it can be unclear \emph{why} a certain pixel or group of pixels were important - color, geometry, texture, etc. could all be reasons why a pixel might have been important. Thus, we observe that incorporating physical models, such as renderers, into learning, as done in our proposed method, provides an avenue for improved interpretability. By leveraging existing gradient-based attribution methods \cite{selvaraju2017grad,sundararajan2017axiomatic}, we can determine how much the features of each physical scene parameter contributed to a prediction, and generate corresponding saliency maps for each feature, as shown in Figure \ref{figure:gradcam}. As differentiable renderers become more realistic and are able to model more scene parameters, interpretability can continue to be improved when renderers are used in training.



\vspace{1mm}
\noindent
\textbf{Disentanglement \& Robustness} We observe that our methods improve \emph{physical} disentanglement and robustness. The learned features are predictive of scene parameters, which themselves are disentangled, as shown in \cite{wu2020unsupervised}, and can be rendered to form an image. We compute Pearson's correlation coefficient (PCC) between each combination of the four learned physical features, across multiple datasets, both with and without our LOOCC loss. Not only do results indicate low correlation between features, but also that LOOCC further reduces correlation. Without LOOCC, our method yields a mean PCC of 0.28 and 0.26 on the Buffy and BBT datasets, respectively, whereas, with LOOCC, mean PCC is 0.18 on both datasets. Disentanglement is also supported empirically by our model's robustness to novel lighting and camera views. Our method, which omits camera and light features for testing, yields the biggest improvement over other methods when evaluated on Buffy and BBT, which contain novel lighting and views. More work is needed to study whether including a renderer in training and disentangling physical features can sufficiently mitigate the susceptibility of deep learning models to misclassifications under novel lighting and views \cite{madan2020and,madan2021small}.

\vspace{1mm}
\noindent
\textbf{Limitations \& Future Work:} A limitation of our work is the inverse renderer used, which constrains us to symmetrical objects and does not account for specular effects. Future work can build on our method to show the utility of inverse rendering and physical disentanglement for perception of more complex scenes and larger images. In addition, our proposed LOOCC loss samples light and view only, but could sample geometry and albedo if these parameters were modeled probabilistically. Finally, by using a differentiable renderer that models other scene parameters, such as haze, atmospheric conditions, etc, robustness to a variety of physical phenomenon may be improved using our method.

\vspace{-2mm}
\section{Conclusion}
\vspace{-2mm}

We present a framework for disentangling representations with regard to physical scene parameters, such as geometry, albedo, light, and camera view, using inverse rendering. We demonstrate improved performance over strong baselines with our learned features across both face and non-face images, and with clustering, linear classification, and segmentation tasks. We also introduce a novel objective called Leave-One-Out, Cycle Contrastive loss (LOOCC) that leads to improved downstream performance. LOOCC helps to disentangle scene parameters by contrasting images with their physically augmented counterparts, generated by a renderer. Finally, we discuss the implications of using differentiable rendering as part of representation learning on areas such as interpretability and robustness to physical phenomena, including novel lighting and camera views.



\subsection{Acknowledgements}
This research was supported by the SMART Contract IARPA Grant \#2021-20111000004. The authors would also like to thank Shangzhe Wu and Ayush Chopra for valuable conversations related to this research.

%
%
\bibliographystyle{splncs04}
\bibliography{egbib}
\end{document}